\documentclass{article} 
\usepackage{iclr2026_conference,times}

\usepackage{amsmath,amsfonts,bm}

\def\eqref#1{equation~\ref{#1}}

\def\1{\bm{1}}

\DeclareMathAlphabet{\mathsfit}{\encodingdefault}{\sfdefault}{m}{sl}
\SetMathAlphabet{\mathsfit}{bold}{\encodingdefault}{\sfdefault}{bx}{n}

\usepackage{hyperref}
\usepackage{url}
\usepackage{amsmath}
\usepackage{booktabs}
\usepackage{graphicx}
\usepackage{wrapfig}
\usepackage{needspace}
\usepackage{xcolor}
\usepackage{tikz}
\usepackage{booktabs}
\usepackage{xcolor}
\usepackage{colortbl}
\usepackage{multirow}
\usepackage{graphicx}
\usepackage{siunitx}
\definecolor{tablehead}{RGB}{237,242,248}
\definecolor{basebg}{RGB}{247,247,247}
\definecolor{scorebg}{RGB}{240,247,255}
\definecolor{rankbg}{RGB}{248,244,255}

\usetikzlibrary{arrows.meta}
\usepackage{microtype}
\definecolor{themenavy}{HTML}{4F587D}
\definecolor{themeorchid}{HTML}{C68DC0}
\definecolor{themesky}{HTML}{C2E0EE}
\definecolor{themelavender}{HTML}{DBC8ED}
\definecolor{themeorchidpale}{HTML}{F1DDEE}
\definecolor{themegold}{HTML}{FFC259}  
\tikzset{
  var/.style={circle,draw=themenavy,line width=0.4pt,minimum size=11pt,inner sep=0pt,font=\scriptsize},
  arrowtip/.style={-{Stealth[length=3pt,width=2.6pt]}},
  gain/.style={arrowtip,line width=0.9pt,draw=themeorchid},
  nogain/.style={arrowtip,line width=0.45pt,draw=black!35},
  hypgain/.style={arrowtip,line width=0.9pt,draw=black},
  hypnogain/.style={arrowtip,line width=0.5pt,draw=black!55,dash pattern=on 1.6pt off 1.2pt},
  edge/.style={arrowtip,line width=0.55pt,draw=themenavy},
  hit/.style={draw=themegold,line width=1.1pt},
  hitarrow/.style={arrowtip,line width=1pt,draw=themegold},
  panel title/.style={font=\scriptsize,align=center,anchor=south,inner sep=0pt},
  note/.style={font=\tiny,anchor=north,inner sep=0pt},
}
\newcommand{\sqbullet}{\begin{tikzpicture}[x=1pt,y=1pt,baseline=-0.4ex]
\fill[black!35] (0.8,-0.8) rectangle (5.0,3.4);
\fill[black!75] (0,0) rectangle (4.2,4.2);
\end{tikzpicture}}
\newcommand{\hypdag}[5]{%
\begin{tikzpicture}[x=1pt,y=1pt]
\node[var,fill=themesky] (x) at (0,0) {$x$};
\node[var,fill=themelavender] (c) at (21,0) {$c$};
\node[var,fill=themeorchidpale] (s) at (42,0) {$s$};
\draw[#3] (x) -- (c);
\draw[#4] (c) -- (s);
\draw[#5] (x.north) to[out=40,in=140,looseness=0.9] (s.north);
\node[panel title] at (21,15) {\textbf{#1}\\#2};
\end{tikzpicture}}

\title{Of Chains and Gains: What Reinforcement Learning changes on Reasoning-based Image Quality Assessment}

\author{%
  Yixuan Li\\
  University of Bristol, UK\\
  \texttt{yixuanli423@gmail.com}
  \And
  Yuxuan Jiang\\
  University of Bristol, UK\\
  \texttt{yuxuan.jiang@bristol.ac.uk}
  \AND
  Hanwei Zhu\\
  Jiangxi University of Finance\\
  and Economics, China\\
  \texttt{zhuhanwei@jxufe.edu.cn}
  \And
  Hancheng Zhu\\
  China University of Mining\\
  and Technology, China\\
  \texttt{zhuhancheng@cumt.edu.cn}
  \AND
  Fan Zhang\\
  University of Bristol, UK\\
  \texttt{fan.zhang@bristol.ac.uk}
  \And
  Shiqi Wang\\
  City University of Hong Kong, Hong Kong\\
  \texttt{shiqwang@cityu.edu.hk}
  \AND
  David Bull\\
  University of Bristol, UK\\
  \texttt{Dave.Bull@bristol.ac.uk}
}

\hypersetup{
  pdftitle={Of Chains and Gains: What Reinforcement Learning changes on Reasoning-based Image Quality Assessment},
  pdfauthor={Yixuan Li, Yuxuan Jiang, Hanwei Zhu, Hancheng Zhu, Fan Zhang, Shiqi Wang, David Bull},
  pdfsubject={arXiv preprint}
}

\iclrfinalcopy 
\begin{document}

\maketitle

\begin{abstract}
Reinforcement Learning with Verifiable Rewards (RLVR) has improved the performance of large Vision-Language Models (VLMs) on image quality assessment (IQA). Existing approaches often attribute these gains to improved reasoning before scoring, supported by increasingly plausible Chain-of-Thought (CoT). However, whether these gains exceed sampling effects and how they depend on explicit reasoning remain unclear. We rigorously investigate these questions through sampling-aware evaluation, controlled reasoning interventions, and shortcut experiments. We find that RL-trained models outperform base models under large-budget $\operatorname{max}_g@k$ evaluation, while improvements beyond averaging are concentrated in-distribution. RLVR produces chains that raise base-model performance when transplanted and improves scoring with the chain held fixed. Yet scoring becomes less dependent on the chain during inference, more clearly at 7B than at 32B. Steering the probed quality direction in thinking tokens also has little effect on scoring. Finally, RL can improve scoring accuracy by exploiting textual cues correlated with mean opinion scores (MOSs), which could inflate benchmark performance. Together, these findings show that RLVR raises quality scoring and produces more useful chains, but that better scoring does not imply a stronger reliance on explicit reasoning, while score-based rewards can also encourage reliance on quality-irrelevant shortcuts. We hope this preliminary study can enlighten more and more concrete reasoning schemes in IQA.

\end{abstract}

\section{Introduction}
\label{sec:intro}
Image quality assessment (IQA) predicts the perceptual quality of images as judged by human observers~\citep{wang2006modern}. Large vision-language models (VLMs) have been widely adapted to this task, initially by supervised fine-tuning~\citep{you2023depicting, wu2023qalign, zhu2024adaptive} and recently by Reinforcement Learning with Verifiable Rewards (RLVR)~\citep{li2025qinsight, wu2025visualquality,jiang2026q}. RLVR-based IQA methods elicit VLMs to reason with structured chain-of-thoughts (CoTs)~\citep{wei2022chain} before scoring\footnote{We restrict our scope to explicit reasoning which forces models to output readable reasoning.}, and reward the agreement between predictions and mean opinion scores (MOSs) automatically. They achieved state-of-the-art quality prediction and generalisation without literally supervising the reasoning process. It is widely believed that RLVR can enable VLMs to autonomously incentivize in-depth image quality reasoning~\citep{li2025qinsight}, such as better awareness of distortion types, degradation levels, and corruption localization, which are evidenced by plausible and comprehensive CoT explanations absent in base models~\citep{guo2025deepseek}. Consequently, RLVR is viewed as a route for evolving VLMs for better IQA capabilities~\citep{cai2025qponder, jia2025refine}, potentially advancing closer towards human visual perception.

Despite these empirical gains, RLVR on IQA remains underexamined. The quality reward verifies only the score, so the training objective does not determine whether an improvement of the score is obtained through the chain. This raises a fundamental question:

\textbf{\textit{Does RLVR genuinely improve the quality reasoning of VLMs, with better chains leading to better scores, or can it improve the performance bypass the chain?}}

Previous belief rests on two kinds of evidence, higher rank correlation and more plausible chains, while neither establishes that the improved performance is causally attributes to reasoning. RLVR is known to sharpen the output distribution of the base model, which is how it raises single-sample accuracy in mathematics and programming~\citep{yue2025does}. RL-trained IQA models do produce more concise chains with less unrelated content~\citep{zhao2026reasoning}, which might be what a sharpened output distribution looks like in the text. On the other hand, plausible chains do not causally show that the score is derived from them, since IQA relies densely on vision~\citep{bosse2018ssvep}, and language chains can fail to reflect the actual cause of the answer even after outcome-based RL~\citep{turpin2023language, lanham2023measuring, chen2025reasoning, yu2026outcome}. Since quality reward only verifies agreement with MOS, any correlate of MOS in the training images raises it as readily as quality itself~\citep{skalse2022defining, geirhos2020shortcut, shao2025spurious}, so higher accuracy with plausible chains cannot tell better reasoning apart from reduced sampling randomness, a path that bypasses the chain, or learned underlying shortcuts.

\begin{figure}[t]
\centering
\includegraphics[width=\linewidth]{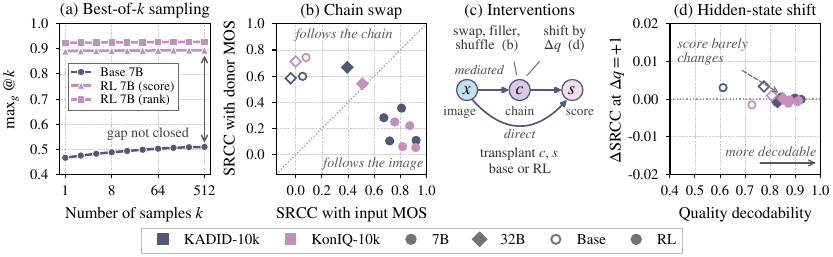}
\vspace{-14pt}
\caption{Gains and chains under RLVR for IQA. (a) $\operatorname{max}_g@k$ SRCC on KADID-10k. (b) Scores under a swapped chain, SRCC with the input image against the donor image. (c) The mediated path through the chain, the direct path, and where the interventions act. (d) Quality decodability of the chain against the change in SRCC when its hidden state is steered.}
\label{fig:teaser}
\vspace{-12pt}
\end{figure}
To investigate this question, we design a set of controlled experiments that address the three alternatives in turn, summarized in Figure~\ref{fig:teaser}. To separate the gain from sampling randomness, we note that one prediction per image conflates what a model can assess with how reliably it samples, and therefore adopt $\operatorname{max}_g@k$~\citep{walder2025passk}, the expected best Spearman rank correlation coefficient (SRCC) among $k$ complete test-set predictions sampled from a model, asking whether a base model reaches its RL-trained counterpart under a larger sampling budget (Figure~\ref{fig:teaser}a). If the gain survives this comparison, the next question is whether it is obtained through the chain. Since the image can influence the score through the chain or directly, and a chain that is useful for scoring need not be one that the score depends on, we fix the image and intervene on the chain in three ways (Figure~\ref{fig:teaser}c). Swapping in the chain of another image or removing its content reveals whether the score depends on the chain (Figure~\ref{fig:teaser}b). Transplanting chains between base and RL-trained models then determines whether the chain nevertheless carries the gain, and shifting the hidden representation of the chain along the direction that encodes quality probes whether the information encoded there is used (Figure~\ref{fig:teaser}d). Finally, we train models with textual cues whose correlation with MOS is controlled, which examines whether accuracy can rise without a better assessment of the image. We discovered several perhaps surprising findings that may help to better understand VLM's reasoning in IQA, as follows:
{\renewcommand{\labelitemi}{\sqbullet}\setlength{\leftmargini}{1.5em}\setlength{\topsep}{2pt}\setlength{\partopsep}{0pt}%
\begin{itemize}\setlength{\itemsep}{1pt}\setlength{\parsep}{0pt}\setlength{\parskip}{0pt}
\item \textbf{RLVR improves quality prediction beyond sampling effects, and the part beyond averaging is concentrated in-distribution.} Under $\operatorname{max}_g@k$, a base model with a large sampling $k$ remains below its RL-trained counterpart, and in-distribution the RL-trained model stays well above the base model even after the sampling noise of the base model is averaged out, whereas out-of-distribution this margin can vanish. This suggests that unlike mathematics or programming tasks, RL \textit{does} improve VLM's image quality scoring capability, but it is confined to in-distribution test sets, whereas out-of-distribution RLVR makes base models more reliable in a single sampling.
\item \textbf{RLVR produces more useful chains on IQA, yet higher performance comes with a weaker dependence on them during inference. } Through chain intervention experiments on both text and hidden state, we discovered RLVR improves both the usefulness of the chain and the scoring model itself. However, this joint improvement does not come with a stronger dependence on the chain. The 7B models show less dependence than the 32B model.
\item \textbf{RLVR can learn shortcuts unrelated to visual quality.} We discover that when a textual cue is related to MOS during training, RL models can gain improvements via predicting this shortcut, even when the cue carries no quality semantics, while models trained with the same text uncorrelated with MOS are unaffected. This indicates that the reward can be easily met by correlate of MOS, so improvements after RLVR may partly reflect such correlates in the data.
\end{itemize}}

In conclusion, we discovered RLVR can actually improve VLM's capability on image quality scoring and produces more useful chains, yet it can indicate a weaker reliance on explicit reasoning during inference than base models. The gain is not carried by the chain alone but also by an improved mapping from image and chain to score. This highlights a gap between using RLVR to achieve accurate and explainable IQA, underscoring more faithful image quality reasoning.
\vspace{-5pt}
\section{Related Works}
\label{sec:related}
\vspace{-6pt}
\paragraph{\textbf{CoT faithfulness and mechanistic interpretability.}}
A faithful chain of thought (CoT) reflects model's decision process rather than only being plausible~\citep{jacovi2020towards}. CoT~\citep{wei2022chain} often falls short of it, since models can omit what drives their answers~\citep{turpin2023language}, the answer often survives a truncated chain~\citep{lanham2023measuring, paul2024making}, and it follows a corrupted step less after post-training for reasoning~\citep{jia2026decorative}. Post-training weakens the link between chain and answer~\citep{bao2025likely}, and outcome-based RL raises accuracy without making the chain faithful~\citep{chen2025reasoning} or causally important~\citep{yu2026outcome,dutta2024think,li2023inference, ju2025probing, zhang2025latent}. These works address tasks in which the chain is essential and correctness is exact. We bring this perspective to the IQA task, where the answer rests on dense visual evidence and has no exact correctness. We further contrast base and RL-trained models to measure how RLVR changes the dependence of the score on the chain.
\vspace{-8pt}
\paragraph{\textbf{RLVR for IQA.}}
VLM-based IQA progressed from benchmarking~\citep{wu2023qbench} to supervised fine-tuning for opened-question~\citep{wu2023q, you2023depicting} and scoring~\citep{wu2023qalign, zhu2024adaptive, you2025teaching}. Recently, reasoning-based methods elicit models to reason before scoring~\citep{li2025qinsight, wu2025visualquality, cai2025qponder, jia2025refine} and obtained significant improvements, which are widely attributed to learned quality reasoning. \citet{zhao2026reasoning} trace the generalization of RL models to the representation formed by textual reasoning, and \citet{li2026pure} discover that quality derived from the text alone fall well below those from the image. These indicate that the reasoning text can suffice to support the score under essential conditions, indicating the sufficiency side of faithfulness~\citep{lyu2023faithful}, but not that the score must pass through it, \textit{i.e.}, the necessity. It still unclear whether quality scores depend on chains, whether the gains persist at large sampling budgets, or whether the reward is met through shortcuts~\citep{geirhos2020shortcut}.
\vspace{-8pt}
\section{Preliminaries}
\paragraph{\textbf{Reinforcement learning with verifiable rewards.}}
\vspace{-6pt}
\label{sec:rlvr}
Let $\pi_\theta$ be a VLM that, given an image $x$ and a scoring prompt, generates a response $y=(c,s)$, a reasoning chain $c$ followed by a quality score $s$. The verifier is the MOS $m$ of $x$. The \emph{score-based} reward~\citep{li2025qinsight} is granted when $s$ falls within a tolerance of $m$, and the \emph{rank-based} reward~\citep{wu2025visualquality} measures how well the preference between two images agrees with the order of MOSs. Refer to Appendix for reward details. Training maximizes $J(\theta)=\mathbb{E}_{x\sim\mathcal{D}}\,\mathbb{E}_{y\sim\pi_\theta(\cdot|x)}[r]$ with GRPO~\citep{shao2024deepseekmath}. In this study, the reasoning content is never supervised explicitly to better isolate the effects.
\vspace{-8pt}
\paragraph{\textbf{Reasoning paradigms for IQA.}}
\label{sec:paradigms}
RLVR-based IQA methods adopt a \emph{think-then-score} paradigm, where the model generates a reasoning chain before the quality score. Omitting the prompt, the response distribution factorizes as $\pi_\theta(c,s\mid x)=\pi_\theta(c\mid x)\,\pi_\theta(s\mid x,c)$. Therefore, the image could influence the score along either a mediated path $x\to c\to s$, a direct path $x\to s$ that bypasses it, or both simultaneously. To separate them, we use \emph{intervened scoring} to manage on $\tilde{c}$ and see its influence on $s$, such as swapping and steering (as detailed in Section~\ref{sec:cot}).
\vspace{-8pt}
\paragraph{Models and benchmarks.}
\label{sec:models}
We use the generally employed Qwen2.5-VL-Instruct~\citep{bai2025qwen25vl} at 7B and 32B as base models. For each recipe we train models with the official implementation on synthetic-distorted KADID-10k~\citep{lin2019kadid}, and on authentic-distorted KonIQ-10k~\citep{hosu2020koniq}, respectively. We use the same prompt structure and data throughout the training, and only reward on \textit{quality scoring} regarding quality-relevant supervision. Refer to Appendix for reward function details. Following the splits of \citet{you2025teaching}, we evaluate all models on the test sets of KADID-10k and KonIQ-10k, and additionally on SPAQ~\citep{fang2020perceptual}, AGIQA-3K~\citep{li2023agiqa} and CSIQ~\citep{larson2010most}, which are unseen throughout. Appendix details the datasets used for training and testing.
\vspace{-6pt}
\section{RLVR's Effect on IQA Reasoning Capacity of VLMs}
\vspace{-5pt}
\subsection{Setups}
\label{sec:setups}
\paragraph{Evaluation protocols.} Every base and RL-trained model is sampled $n=512$ times per image\footnote{We employ vLLM scheme~\citep{kwon2023efficient} to ensure independent and identically distributed sampling.}, with temperature 1.0, top-$p$ 0.95 and at most 4,096 generated tokens. All models are tested zero-shot with the same prompt as used in training to avoid any confounding effects. We extract the score enclosed by answer tags. A response without a parsable score is not resampled and counts as an invalid draw within the budget. Surprisingly, although without RLVR, the valid-draw rate can reach 99.5\% for base models, and nearly 100\% for RL-trained models. Appendix lists the prompts.
\vspace{-8pt}
\paragraph{Metrics.} Measuring the reasoning ability is challenging, since methods like single-draw greedy decoding or multi-draw averaging can introduce biases~\citep{wang2023selfconsistency, holtzman2019curious}. Generally, pass@$k$~\citep{chen2021evaluating} is employed to evaluate model's reasoning boundary~\citep{yue2025does}, measuring whether at least one of $k$ responses passes a binary verifier. We employ the continuous extension of pass@$k$, \textit{i.e.}, $\operatorname{max}_g@k$~\citep{walder2025passk} to evaluate rank correlation performance in IQA. Each candidate $X_j$, for $j=1,\ldots,n$, is a complete test-set prediction with one sampled score per image. Taking $m$ for the vector of MOSs over the test set, we define the reward function as $g(X)=\operatorname{SRCC}(X,m)$, evaluated on the images of $X$. MOSs are used only to select among complete candidates, never among the scores of an individual image. Thus $\operatorname{max}_g@k$ is the expected best SRCC among $k$ independently sampled candidates $X'_1,\ldots,X'_k$. We compute it from the $n$ observed rewards $g_j=g(X_j)$, sorted as $g_{(1)}\le\cdots\le g_{(n)}$, with the unbiased estimation,
\begin{equation}
\operatorname{max}_g@k
:=\mathbb{E}_{X'_1,\ldots,X'_k}\left[\max_{1\le j\le k} g(X'_j)\right]
\approx\sum_{i=k}^{n}\frac{\binom{i-1}{k-1}}{\binom{n}{k}}\,g_{(i)},
\label{eq:maxg}
\end{equation}
for $k\in\{1,2,4,\ldots,n\}$. The metric is non-decreasing in $k$ and reduces to pass@$k$ for binary rewards, and its estimate equals the mean candidate SRCC at $k=1$ and the largest observed SRCC at $k=n$. With this measure, we can easily estimate $\operatorname{max}_g@k$ with low variance. Appendix lists more details.
\subsection{IQA Reasoning Capacity Results}
\vspace{-3pt}
\label{sec:results_sampling}
\begin{figure}[t]
\centering
\includegraphics[width=\linewidth]{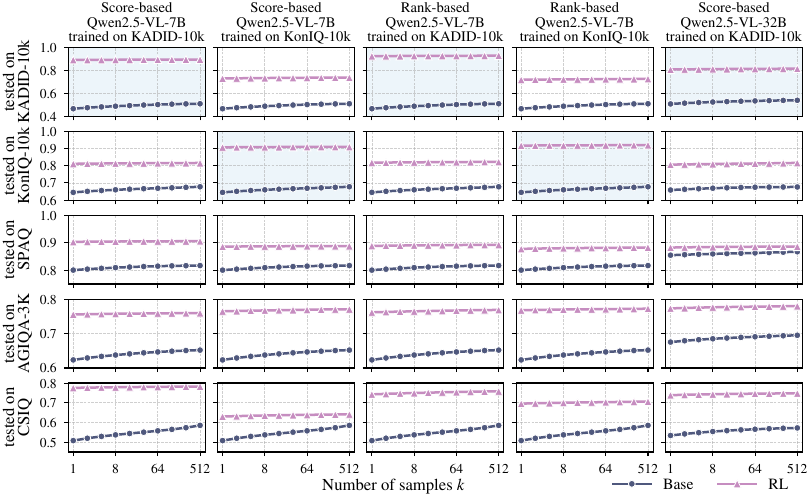}
\vspace{-20pt}
\caption{$\operatorname{max}_g@k$ curves for base and RL models. Shaded panels are in-distribution test sets.}
\label{fig:maxofk}
\vspace{-8pt}
\end{figure}
\begin{figure}[t!]
\centering
\includegraphics[width=\linewidth]{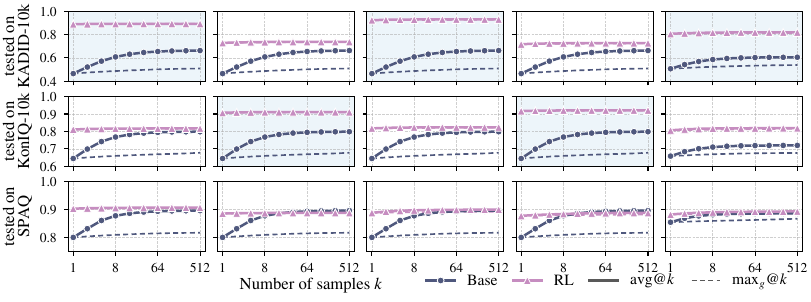}
\vspace{-20pt}
\caption{$\operatorname{avg}@k$ curves for base and RL models. Columns follow Figure~\ref{fig:maxofk}. Refer to Appendix for results on all test sets.}
\label{fig:meank_main}
\vspace{-12pt}
\end{figure}

\paragraph{Effect of RLVR: boosted quality scoring capability at all sampling budgets.}
Figure~\ref{fig:maxofk} shows a consistent ordering across sampling budgets, with every RL-trained model above its base model on every test set and at both scales. At $k=1$, the average-case SRCC of a single draw, RL-trained models lead by 0.03 to 0.46 SRCC, in line with the improvements reported for both recipes~\citep{li2025qinsight,wu2025visualquality}, so RLVR makes a single sampled score considerably more accurate. The improvement persists as $k$ grows to the hundreds. Base models gain more from additional sampling, yet it only closes a part of the gap with RL-trained models. At $k=512$, the difference still remains positive in every comparison, and even the best of 512 base candidates lies below a single RL draw. Therefore, RLVR does raise the quality scoring capability beyond sampling effect.
\vspace{-10pt}
\paragraph{Nature of the gain: denoising rather than coverage, better mapping concentrated in-distribution.}
Interestingly, this discovery looks contrast with the crossover that \citet{yue2025does} report on mathematics, code and visual reasoning tasks, where base models overtake RL-trained ones at large $k$. This indicates that, in a discrete open answer space, a large budget uncovers rarely sampled solutions, while IQA's SRCC aggregates over every image, the random error of individual scores moves it little from one candidate to the next, which therefore concentrate. RLVR sharpens this distribution by removing noise around the score of each image rather than alternative solutions (Appendix contains detailed statistics). To investigate how much of the gain is denoising alone, we further evaluate $\operatorname{avg}@k$, the SRCC of the per-image average of $k$ scores. As shown in Figure~\ref{fig:meank_main}, averaging lifts base models more and saturates quickly. This indicates the single-draw gain of RLVR contains a denoising part, which the base model recovers by averaging its own draws, and a mapping part, which it can not. The mapping part is large in-distribution but small or absent out-of-distribution, where the averaged base model often reaches or crossovers a single RL draw and the crossover of \citet{yue2025does} reappears. Therefore, RLVR improves the quality scoring in-distribution, and the generalization that \citet{zhao2026reasoning} report thus is inherently contained in the base model.

\vspace{-12pt}
\begin{wrapfigure}{r}{0.45\linewidth}
\vspace{0pt}
\centering
\includegraphics[width=\linewidth]{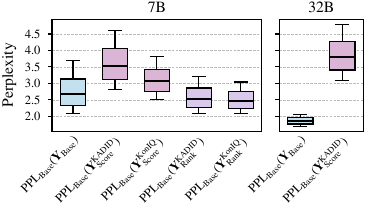}
\vspace{-16pt}
\caption{Perplexity of sampled responses under the base model of the same scale.}
\label{fig:ppl}
\vspace{-9pt}
\end{wrapfigure}
\paragraph{Perplexity analysis.}
To see whether RL-trained responses remain likely under base models, we computed perplexity~\citep{jelinek1977perplexity} of responses from each model, which measures the model’s ability to predict the given response conditioned on the inputs. Lower perplexity indicates that the model has a higher likelihood of generating this response. We randomly sample 16 reasoning for each image, denoted $\mathbf{Y}_{\text{Base}}$, $\mathbf{Y}_{\text{Score}}$ and $\mathbf{Y}_{\text{Rank}}$ by the reward of the model, score every response under the base model of the same scale, and Figure~\ref{fig:ppl} averages the test sets (Appendix gives details). $\operatorname{PPL}_{\text{Base}}(\mathbf{Y}_{\text{Rank}}\mid x)$ occupies the lower part of the distribution of $\mathbf{Y}_{\text{Base}}$, so rank rewards concentrate sampling on the responses that the base model most readily produces. $\operatorname{PPL}_{\text{Base}}(\mathbf{Y}_{\text{Score}}\mid x)$ shifts upward: at 7B these responses are less typical of the base model but still partly within its range, whereas at 32B they lie almost entirely above the narrow band of the base model. Score rewards thus move the text further outside the base distribution as scale grows, although neither reward targets the reasoning text, and in both cases the scores change substantially. This raises the question: \textit{whether the gain passes through this altered textual CoT or bypasses it?}

\vspace{-8pt}
\section{Causal Role of CoT in IQA}
\vspace{-6pt}
\label{sec:cot}
\subsection{Setups}
\begin{wrapfigure}{r}{0.44\linewidth}
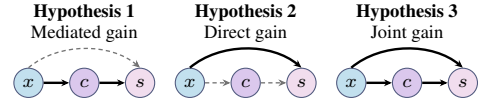

\vspace{-8pt}
\centering
\makebox[\linewidth]{\hypdag{Hypothesis 1}{Mediated gain}{hypgain}{hypgain}{hypnogain}\hfill\hypdag{Hypothesis 2}{Direct gain}{hypnogain}{hypnogain}{hypgain}\hfill\hypdag{Hypothesis 3}{Joint gain}{hypgain}{hypgain}{hypgain}}
\caption{Three hypotheses on source of quality scoring gains, with the image $x$, the chain $c$ and the score $s$. Solid black edges carry the gain and dashed grey edges do not.}
\label{fig:hypotheses}
\vspace{-8pt}
\end{wrapfigure}
\vspace{-6pt}
\paragraph{Hypotheses.} To answer the question above, we state three hypotheses shown in Figure~\ref{fig:hypotheses}. Under H-1 mediated gain, RLVR improves the quality reasoning expressed in the chain and the score is derived from the chain, so the gain is obtained along $x\to c\to s$. Under H-2 direct gain, RLVR improves the mapping directly from image to score along $x\to s$, and the chain is a by-product of the required format that does not carry the gain. Under H-3 joint gain, both paths are strengthened, so the chain becomes more informative about quality, the score becomes more sensitive to the image, and the gain draws on both.

\begin{wrapfigure}[8]{r}{0.50\textwidth}
\vspace{-14pt}
\centering
\makebox[\linewidth][r]{%
\begin{tikzpicture}[x=1pt,y=1pt,var/.append style={font=\footnotesize},note/.style={font=\scriptsize,anchor=north,inner sep=0pt},panel title/.style={font=\scriptsize\bfseries,align=center,anchor=south,inner sep=0pt}]
\node[var,fill=themesky] (x) at (0,0) {$x$};
\node[var,fill=themelavender,hit,dashed] (c) at (21,0) {$\tilde c$};
\node[var,fill=themeorchidpale] (s) at (42,0) {$s$};
\draw[edge] (x) -- (c);
\draw[edge] (c) -- (s);
\node[note] at (21,-7) {swap, filler, shuffle};
\node[panel title] at (21,15) {Chain\\perturbation};
\end{tikzpicture}\hspace{7pt}
\begin{tikzpicture}[x=1pt,y=1pt,var/.append style={font=\footnotesize},note/.style={font=\scriptsize,anchor=north,inner sep=0pt},panel title/.style={font=\scriptsize\bfseries,align=center,anchor=south,inner sep=0pt}]
\node[var,fill=themesky] (x) at (0,0) {$x$};
\node[var,fill=themelavender,hit,dashed] (c) at (21,0) {$c$};
\node[var,fill=themeorchidpale,hit] (s) at (42,0) {$s$};
\draw[edge] (x) -- (c);
\draw[edge] (c) -- (s);
\node[note] at (21,-7) {base $\leftrightarrow$ RL};
\node[panel title] at (21,15) {Cross-model\\transplant};
\end{tikzpicture}\hspace{7pt}
\begin{tikzpicture}[x=1pt,y=1pt,var/.append style={font=\footnotesize},note/.style={font=\scriptsize,anchor=north,inner sep=0pt},panel title/.style={font=\scriptsize\bfseries,align=center,anchor=south,inner sep=0pt}]
\node[var,fill=themesky] (x) at (0,0) {$x$};
\node[var,fill=themelavender,hit] (c) at (21,0) {$c$};
\node[var,fill=themeorchidpale] (s) at (42,0) {$s$};
\draw[edge] (x) -- (c);
\draw[edge] (c) -- (s);
\draw[hitarrow] (21,13) -- (c.north);
\node[font=\footnotesize,anchor=west,inner sep=1pt] at (22,10) {$\Delta q$};
\node[note] at (21,-7) {probe direction};
\node[panel title] at (21,15) {Hidden-state\\perturbation};
\end{tikzpicture}}
\caption{Three kinds of chain interventions. Gold marks where an intervention acts, a dashed outline marks a chain that is replaced.}
\label{fig:interventions}
\vspace{-2pt}
\end{wrapfigure}
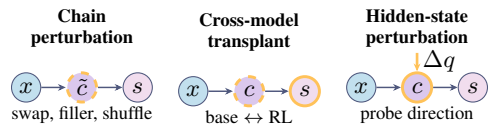
\vspace{-6pt}
\paragraph{Methods.} In order to probe the hypotheses, we fix the input image, original prompt, and only manage on the chain closed by \texttt{\textless /think\textgreater} that precedes the score. Then we let the models to generate the complete answer, from which the score is parsed from the \texttt{\textless answer\textgreater} tags. The first two act on the text of the CoT, and the third on its hidden state, detailed as follows:
{\renewcommand{\labelitemi}{\sqbullet}\setlength{\leftmargini}{1.5em}\setlength{\topsep}{2pt}\setlength{\partopsep}{0pt}%
\begin{itemize}\setlength{\itemsep}{1pt}\setlength{\parsep}{0pt}\setlength{\parskip}{0pt}
\item \textbf{Chain perturbation} replaces the chain with the one that the same model generated for a randomly drawn donor image (\emph{swap}), with filler tokens of the same length (\emph{filler}), or with its own words in random order (\emph{shuffle}). \textit{Swap} describes another image, and thus measures how strongly the score follows the mediated path. \textit{filler} and \textit{shuffle} remove the content of the chain while preserving length and format, so they test whether the content is necessary for scoring.
\item \textbf{Cross-model chain transplant} tests whether a better CoT carries the gain. We only exchange chains sampled for the same image. Each RL-trained model generates quality scores with the chains of its base model, and vice versa. With the model fixed, the gap between the two chains is the gain carried by CoTs; and with the chain fixed, the difference between the two models is the gain that the chain does not carry. H-1 predicts the former, H-2 the latter, and H-3 both.
\item \textbf{Hidden-state perturbation} acts on CoT's hidden states to see if we can steer the score. We fit a continuous linear probe of MOS on the hidden state of the last thinking token before \texttt{\textless/think\textgreater} for every layer, shift this hidden state along the probe direction by the smallest displacement that changes the probe reading by a dose $\Delta q\in\{\pm0.25,\pm0.5,\pm1\}$ on the $[1,5]$ scale, and generate one answer per dose. A random direction orthogonal to the probe direction, with the same norm, serves as the orthogonal control. Details for probe training and steering can be found in Appendix.
\end{itemize}}
We evaluate the models of Section~\ref{sec:models} on KADID-10k and KonIQ-10k with the sampling parameters of Section~\ref{sec:setups}, and apply the three interventions. The unmodified chains form the reference condition. Appendix gives the full protocol.
\vspace{-10pt}
\paragraph{Metrics.} Let $s$ and $s'$ be the quality scores of the test images after uniform reasoning and under an intervention, and $m$ their MOSs. Under \textit{swap}, we report two correlations: $\rho=\operatorname{SRCC}(s',m)$ with the MOSs of the input images, which is high when the score follows the image, and $\gamma=\operatorname{SRCC}(s',m_z)$ with the MOSs $m_z$ of the donor images, which is high when the score follows the chain. We further report $\Delta\operatorname{SRCC}=\rho-\operatorname{SRCC}(s,m)$ to measure the influence of intervention on each dataset. Perturbation success rate (PSR)~\citep{ju2025probing} is adapted to measure the influence on individual image, which measures the fraction of images whose score moves towards the intended direction, \textit{i.e.}, the quality of the donor image under \textit{swap}, and the sign of $\Delta q$ under \textit{hidden-state perturbation}. \textit{Filler} and \textit{shuffle} define no intended direction, so PSR is not computed. 
\vspace{-4pt}

\definecolor{indomainbg}{RGB}{237,245,250} 

\begin{table}[t]
\caption{\textbf{CoT intervention results.}
Ref.\ denotes SRCC with the original chains. Refer to \textit{\textbf{Metrics}} for details of employed measure. Transplant results are reported as model/chain: RL/Base uses
the RL-trained model to score with base-model chains, while Base/RL denotes the reverse.}
\label{tab:interventions}
\centering
\fontsize{8}{9}\selectfont
\renewcommand{\arraystretch}{1}
\setlength{\tabcolsep}{3.5pt}
\sisetup{detect-weight=true,detect-family=true,table-number-alignment=center}

\begin{tabular}{@{}c l
S[table-format=1.3]
S[table-format=-1.3]
S[table-format=-1.3]
S[table-format=1.3]
S[table-format=1.3]
S[table-format=-1.3]
S[table-format=-1.3]
S[table-format=1.3]
S[table-format=1.3]@{}}

\toprule
\multirow[c]{2}{*}{Test set} &
\multicolumn{1}{c}{\multirow[c]{2}{*}{Model}} &
\multicolumn{1}{c}{\multirow[c]{2}{*}{\textit{Ref}.}} &
\multicolumn{4}{c}{\textbf{Swap} ($\ \Delta$SRCC $=\rho\,-\,$\textit{Ref}.)} &
\multicolumn{1}{c}{\textbf{Filler}} &
\multicolumn{1}{c}{\textbf{Shuffle}} &
\multicolumn{2}{c}{\textbf{Transplant SRCC}} \\
\cmidrule(lr){4-7}\cmidrule(lr){8-8}\cmidrule(lr){9-9}\cmidrule(lr){10-11}
& & &
\multicolumn{1}{c}{$\rho$} &
\multicolumn{1}{c}{$\Delta$SRCC} &
\multicolumn{1}{c}{$\gamma$} &
\multicolumn{1}{c}{\textit{PSR}} &
\multicolumn{1}{c}{$\Delta$SRCC} &
\multicolumn{1}{c}{$\Delta$SRCC} &
\multicolumn{1}{c}{RL/Base} &
\multicolumn{1}{c}{Base/RL} \\
\midrule

& Base 7B  & 0.601 & 0.053 & -0.548 & 0.598 & 0.707 & 0.050 & 0.019 &
\multicolumn{1}{c}{--} & \multicolumn{1}{c}{--} \\
& Base 32B & 0.576 & -0.038 & -0.614 & 0.584 & 0.709 & 0.017 & 0.013 &
\multicolumn{1}{c}{--} & \multicolumn{1}{c}{--} \\
\addlinespace[1.5pt]

&
\multicolumn{10}{>{\columncolor{indomainbg}[\tabcolsep][2\tabcolsep]}l}{
\hspace{0.25em}\textbf{KADID-10k training set} \textit{(in-distribution)}
} \\[-1pt]
\rowcolor{indomainbg}\cellcolor{white} &
\hspace{1em}Score 7B & 0.893 & 0.809 & -0.084 & 0.355 & 0.851 & 0.000 & -0.001 & 0.879 & 0.876 \\
\rowcolor{indomainbg}\cellcolor{white} &
\hspace{1em}Score 32B & 0.819 & 0.393 & -0.426 & 0.667 & 0.806 & -0.013 & -0.006 & 0.659 & 0.792 \\
\rowcolor{indomainbg}\cellcolor{white} &
\hspace{1em}Rank 7B & 0.931 & 0.920 & -0.011 & 0.108 & 0.743 & -0.001 & -0.001 & 0.909 & 0.861 \\

\addlinespace[1pt]

&
\multicolumn{10}{l}{
\hspace{0.25em}\textbf{KonIQ-10k training set} \textit{(out-of-distribution)}
} \\[-1pt]
&
\hspace{1em}Score 7B & 0.739 & 0.674 & -0.065 & 0.282 & 0.755 & -0.002 & 0.000 & 0.718 & 0.712 \\
\multirow{-9}{*}{\rotatebox[origin=c]{-90}{\textbf{KADID-10k}}} &
\hspace{1em}Rank 7B & 0.729 & 0.719 & -0.010 & 0.104 & 0.724 & -0.004 & 0.001 & 0.729 & 0.675 \\

\midrule

& Base 7B  & 0.761 & 0.078 & -0.683 & 0.743 & 0.787 & 0.018 & 0.001 &
\multicolumn{1}{c}{--} & \multicolumn{1}{c}{--} \\
& Base 32B & 0.709 & -0.001 & -0.710 & 0.712 & 0.762 & -0.005 & 0.004 &
\multicolumn{1}{c}{--} & \multicolumn{1}{c}{--} \\
\addlinespace[1.5pt]

&
\multicolumn{10}{>{\columncolor{indomainbg}[\tabcolsep][2\tabcolsep]}l}{
\hspace{0.25em}\textbf{KonIQ-10k training set} \textit{(in-distribution)}
} \\[-1pt]
\rowcolor{indomainbg}\cellcolor{white} &
\hspace{1em}Score 7B & 0.910 & 0.871 & -0.039 & 0.221 & 0.848 & 0.000 & 0.000 & 0.872 & 0.893 \\
\rowcolor{indomainbg}\cellcolor{white} &
\hspace{1em}Rank 7B & 0.921 & 0.917 & -0.004 & 0.054 & 0.821 & 0.000 & 0.000 & 0.912 & 0.867 \\

\addlinespace[1pt]

&
\multicolumn{10}{l}{
\hspace{0.25em}\textbf{KADID-10k training set} \textit{(out-of-distribution)}
} \\[-1pt]
&
\hspace{1em}Score 7B & 0.818 & 0.757 & -0.061 & 0.250 & 0.813 & -0.004 & -0.002 & 0.818 & 0.833 \\
&
\hspace{1em}Score 32B & 0.819 & 0.509 & -0.310 & 0.542 & 0.810 & -0.001 & -0.001 & 0.747 & 0.792 \\
\multirow{-9}{*}{\rotatebox[origin=c]{-90}{\textbf{KonIQ-10k}}} &
\hspace{1em}Rank 7B & 0.825 & 0.817 & -0.008 & 0.061 & 0.730 & -0.002 & -0.002 & 0.829 & 0.832 \\

\bottomrule
\end{tabular}
\vspace{-10pt}
\end{table}
\subsection{Reasoning Perturbation Results}
\begin{figure}[t]
\centering
\includegraphics[width=\linewidth]{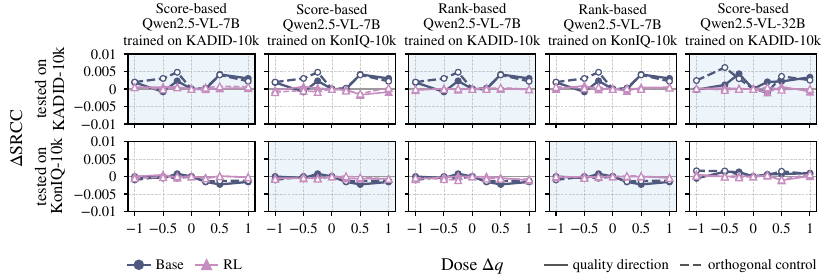}
\vspace{-12pt}
\caption{CoT hidden-state perturbation results against $\Delta q$. Solid lines shift the hidden state of the last thinking token at every layer. Dashed lines apply the orthogonal control of equal norm. Rows are the test sets whose training splits fitted the probes, and shaded panels are in-distribution test sets.}
\vspace{-10pt}
\label{fig:hidden_perturb}
\end{figure}

\vspace{-4pt}
\paragraph{Chains have less effect on quality scoring performance after RLVR.} We first test H-1, under which the score is derived from the chain, with three kinds of perturbations. As shown in Table~\ref{tab:interventions}, after \textit{swapping}, all PSRs are non-zero, so every model responds to the swapped chain. Base model's prediction is more correlated with donor's MOS instead of the input image's, indicated by $\gamma$ is closer to Ref. than $\rho$. In contrast, 7B RL models show an opposite trend, where quality scores follows the input image more than an irrelevant CoT, while 32B RL models lies between the two. This demonstrates that RLVR moves the score from the mediated to the direct path, almost completely at 7B and partially at 32B. Meanwhile, removing the content of the chain gives a slightly different picture: \textit{filler} tokens of the same length or \textit{shuffling} the chain's own words can barely change model's performance for both the base and RL models. As such, the gain does not only rely on the model to derive the score from an explicit chain, while larger models (32B) can be more reliant. Therefore, H-1 does not explain it. \citet{zhao2026reasoning} does not conflict with us, where scores do attend to the chain without being determined by its content. \textit{It is worth noting that this finding concerns the use of the chain at inference and not its role during training.}
\vspace{-8pt}
\paragraph{Chains of RL-trained models nevertheless carry quality information for scoring, and scale matters.}
We then examine whether chains contribute, which H-2 denies. As shown in the transplant columns in Table~\ref{tab:interventions}, base models performance can be enhanced by replacing its CoTs with those of its RL-trained counterpart. The chains of the RL-trained models thus carry quality information that the base model can use, in line with their higher quality decodability (Appendix). Meanwhile, with the chains of the base model fixed, replacing the base model by the RL model generally raises performance. Therefore, both CoT and direct mapping from image carry the gain, supporting H-3. Scale Influences How the Two Factors Combine. At 7B the RL-trained model scores almost as well with the chains of its base model as with its own, so the better chain and the better model largely overlap as routes to the same gain, and the score depends little on the chain. At 32B the chain carries most of the gain and the RL-trained model loses part of it without its own chains, so the chain still feeds the score, as under the swap. What we find interesting is that this scale difference parallels the perplexity in Figure~\ref{fig:ppl}, in which the chains distribution of the 32B RL model lifts drastically compared with 7B. This could possibly indicate that the larger the shift of the chain distribution from that of the base model, the more the model relies on the chain.
\vspace{-8pt}
\paragraph{More useful chains do not mean more dependence on chains.}
We further test H-1 at the level of the representation. We find out that RLVR raises the quality decodability of the last thinking token at every layer (Appendix), so if the score were read from this representation, steering it along the probed quality direction would steer the score. However, Figure~\ref{fig:hidden_perturb} shows that scoring can hardly be steered by disturbing CoT's hidden state. Shifting the hidden state by up to one point of the probe reading at every layer changes the SRCC by at most 0.005, no more than the orthogonal control, and base and RL-trained models behave alike. This again shows that the representation is informative about quality, not that the score depends on it, further supporting H-3.
\vspace{-8pt}
\paragraph{Summary.} In conclusion, RLVR improves both chain's usefulness and the scoring model itself, yet the better scores do not come with a stronger dependence on the chain during inference. Since the quality-relevant reward checks the agreement between predictions and MOSs, cues that correlate with MOS can probably be reinforced just as well. The next section therefore tries to answer: \textit{whether RLVR can learn quality-irrelevant shortcuts to a higher scoring performance}.


\vspace{-8pt}
\section{Shortcuts to the Quality Reward}
\label{sec:shortcuts}
\vspace{-6pt}
\subsection{Setups}
\label{sec:shortcut_setups}
\vspace{-4pt}
\paragraph{Methods.} We test whether RLVR learns to exploit prompt cues that are unrelated to image quality. The images and their MOSs stay unchanged, and only a textual cue is appended to the scoring prompt. We use four kinds of cues, each with two variants, noted as P and Q: (i) \emph{A/B identifiers} arbitrarily append an A or B; (ii) \emph{source descriptions} state that the image comes from a good-quality source or a bad one; (iii) \emph{numerical hints} give an unverified preliminary rating of 4.50 or 1.50; and (iv) \emph{irrelevant text length} repeats a neutral sentence eight times or once to change text length. (ii) and (iii) state quality explicitly, whereas (i) and (iv) carry no quality information. For each kind of cue, we train two Qwen2.5-VL-7B-Instruct~\citep{bai2025qwen25vl} models on KADID-10k~\citep{lin2019kadid}, using GRPO with the score-based and format rewards of Section~\ref{sec:rlvr}. We designed two training modes, which use the same images, image order and settings and differ only in whether the cue predicts MOS. Comparing the two trained models therefore shows what RLVR learns from a cue that predicts MOS. In \emph{independent} training, which serves as the control, the cues are shuffled across images and are thus unrelated to MOS. In \emph{correlated} training, P is paired with high-MOS and Q with low-MOS images on approximately 90\% of training images. At test time, we change the relation between cue and MOS. Both trained models and the base model are evaluated on the KADID-10k test split with 16 draws per image under four cue assignments: \emph{aligned} pairs P with high-MOS images as in correlated training, \emph{independent} shuffles the cues, \emph{reversed} pairs P with low-MOS images, and \emph{removed} omits the cue. If the correlated model outperforms its independent control under alignment but falls below it under reversal, it has learned to exploit the cue, and its advantage under alignment cannot be attributed to a better assessment of image quality. Appendix gives the full setting details.
\vspace{-10pt}
\paragraph{Metrics.} Let $\rho_{u,v}$ be the SRCC between the mean valid rating and MOS of the correlated or independent model $u\in\{\mathrm{corr},\mathrm{indep}\}$ under test assignment $v$. We measure how much the correlated model relies on the cue by the interaction
\begin{equation}
I=\big(\rho_{\mathrm{corr},\mathrm{aligned}}-\rho_{\mathrm{indep},\mathrm{aligned}}\big)
 -\big(\rho_{\mathrm{corr},\mathrm{reversed}}-\rho_{\mathrm{indep},\mathrm{reversed}}\big).
\label{eq:shortcut_interaction}
\end{equation}
An improved assessment persists under reversal and gives $I\approx0$, whereas reliance on the cue gives $I>0$. On individual images, the same-image score shift of a model is $\delta=N^{-1}\sum_i[\bar{s}(x_i,P)-\bar{s}(x_i,Q)]$, where $\bar{s}(x_i,P)$ is its mean valid rating of image $x_i$ under cue P and $N$ the number of images with a valid rating under both cues. The pixels are fixed, so $\delta$ is attributable to the cue alone. Appendix details the parsing rule and the treatment of invalid ratings.
\vspace{-5pt}
\subsection{Shortcut Results}
\label{sec:shortcut_results}

\definecolor{corrbg}{RGB}{230,230,230}
\definecolor{indepbg}{RGB}{246,246,246}

\paragraph{Correlated training yields cue-dependent accuracy.}
We first examine whether a cue that predicts MOS during training raises test accuracy, and whether the resulting gain depends on the cue. \begin{wraptable}{r}{0.55\linewidth}
\vspace{-10pt}
\caption{\textbf{Shortcut results.} Corr.\ and Indep.\ denote
models trained with correlated and independent cue assignments. Bold denotes the highest SRCC within each cue and test condition.}
\label{tab:shortcut_results}
\vspace{5pt}
\centering
\fontsize{8}{9}\selectfont
\renewcommand{\arraystretch}{1}
\setlength{\tabcolsep}{2.0pt}
\sisetup{detect-weight=true,detect-family=true,table-number-alignment=center}

\begin{tabular}{@{}l l
S[table-format=1.3]
S[table-format=1.3]
S[table-format=-1.3]
S[table-format=1.3]
S[table-format=-1.3]@{}}

\toprule
\multicolumn{1}{c}{\multirow[c]{2}{*}{Cue}} &
\multicolumn{1}{c}{\multirow[c]{2}{*}{Model}} &
\multicolumn{4}{c}{\textbf{Test assignment: SRCC $\uparrow$}} &
\multicolumn{1}{c}{\multirow[c]{2}{*}{$\delta$}} \\
\cmidrule(lr){3-6}
& &
\multicolumn{1}{c}{Aligned} &
\multicolumn{1}{c}{Independent} &
\multicolumn{1}{c}{Reversed} &
\multicolumn{1}{c}{Removed} & \\
\midrule

& Base   & 0.611 & 0.619 & 0.618 & 0.613 & -0.013 \\
\rowcolor{corrbg}\cellcolor{white} &
Corr.   & {\bfseries 0.843} & {\bfseries 0.824} & 0.799 & {\bfseries 0.821} & 0.108 \\
\rowcolor{indepbg}
\cellcolor{white}\multirow{-3}{*}{(i)} &
Indep.  & 0.818 & 0.819 & {\phantom{$-$}\bfseries 0.819} & 0.818 & -0.004 \\

\addlinespace[1.2pt]

& Base   & 0.727 & 0.571 & 0.407 & 0.613 & 0.235 \\
\rowcolor{corrbg}\cellcolor{white} &
Corr.   & {\bfseries 0.881} & 0.550 & -0.027 & 0.781 & 1.360 \\
\rowcolor{indepbg}
\cellcolor{white}\multirow{-3}{*}{(ii)} &
Indep.  & 0.823 & {\bfseries 0.821} & {\phantom{$-$}\bfseries 0.819} & {\bfseries 0.821} & 0.006 \\

\addlinespace[1.2pt]

& Base   & 0.823 & 0.352 & -0.315 & 0.613 & 0.820 \\
\rowcolor{corrbg}\cellcolor{white} &
Corr.   & {\bfseries 0.878} & 0.521 & -0.011 & 0.760 & 1.372 \\
\rowcolor{indepbg}
\cellcolor{white}\multirow{-3}{*}{(iii)} &
Indep.  & 0.814 & {\bfseries 0.814} & {\phantom{$-$}\bfseries 0.817} & {\bfseries 0.820} & 0.007 \\

\addlinespace[1.2pt]

& Base   & 0.591 & 0.615 & 0.637 & 0.613 & -0.034 \\
\rowcolor{corrbg}\cellcolor{white} &
Corr.   & {\bfseries 0.891} & 0.771 & 0.594 & {\bfseries 0.827} & 0.587 \\
\rowcolor{indepbg}
\cellcolor{white}\multirow{-3}{*}{(iv)} &
Indep.  & 0.822 & {\bfseries 0.824} & {\phantom{$-$}\bfseries 0.825} & 0.823 & -0.009 \\

\bottomrule
\end{tabular}
\vspace{-8pt}
\end{wraptable}
Table~\ref{tab:shortcut_results} shows that correlated training improves aligned-test SRCC over independent training for all four cues, by 0.026 to 0.069. However, under reversed assignments, every correlated model falls below its independent control. The interactions $I$ are 0.046 for A/B identifiers, 0.903 for source descriptions, 0.892 for numerical hints and 0.300 for text length. The independent controls remain between 0.814 and 0.825 SRCC across all cues and test conditions, and all comparisons retain the full test set, with valid-draw rates of at least 99.98\% for the trained models. The advantage under alignment is therefore attributable to the cue and not to a better assessment of the image.
\vspace{-8pt}
\paragraph{Cues with and without quality semantics are both exploited.}
We next examine how strongly the scores follow each cue. For (ii) source descriptions and (iii) numerical hints, which state quality directly, reversing the cues lowers the SRCC of the correlated models from 0.881 and 0.878 to $-0.027$ and $-0.011$. On the same image, replacing Q with P raises their mean rating. The independent models shift by less than 0.01 under the same replacement. Cues that carry no quality judgment are exploited as well, though more weakly. An arbitrary A/B identifier lowers the SRCC, and an irrelevant repeated sentence lowers it significantly with $\delta=0.587$, while removing either cue returns the correlated model to the level of its control. Surface features of the prompt are thus exploited whenever they correlate with MOS, and the more so when they also state quality.
\vspace{-8pt}
\paragraph{Cue sensitivity of the initial model.}
Finally, we examine whether this dependence is acquired through RL or is already present before training. As listed in Table~\ref{tab:shortcut_results}, base models respond to (ii) source descriptions and (iii) numerical hints and its performance under reversed numerical hints is below the correlated model. The effect of RL therefore depends on RL training. Independent training largely removes this sensitivity, reducing both shifts below 0.01, whereas correlated training strengthens them above 1.3 and gains an advantage whenever the cue predicts MOS.

\vspace{-6pt}
\section{Conclusion}
\label{sec:conclusion}
\vspace{-5pt}
This work separates the gains of RLVR in IQA from sampling effects and examines whether they are obtained through explicit reasoning. RL-trained models remain above their base models at every sampling budget, so the gain is not a sampling effect. Averaging the draws of a base model recovers the denoising part of this gain, whereas the improved image-to-score mapping is concentrated in-distribution and small or absent out-of-distribution. Chain interventions on both text and hidden state show that RLVR yields more useful chains and a better scorer at the same time, so both the chain and the direct path from image to score carry the gain. Yet neither improvement brings a stronger dependence on the chain at inference: the scores of RL-trained models follow the input image more than a swapped chain, barely change when the content of the chain is removed, and can hardly be steered along the quality direction of its hidden state, more so at 7B than at 32B. Shortcut experiments further show that the reward is met as readily by textual cues correlated with MOS as by quality itself, even by cues without quality semantics, and that the advantage reverses with the correlation. Higher accuracy and more useful chains therefore do not establish that scoring relies on explicit quality reasoning, and our interventions concern the use of the chain at inference rather than its role during training. Reliable and explainable IQA calls for evaluating chain dependence and robustness to shortcuts, alongside rewards that connect the reasoning to the final judgment.

\bibliography{iclr2026_conference}
\bibliographystyle{iclr2026_conference}

\end{document}